\documentclass[a4paper,fleqn]{cas-dc}
\usepackage{listings}
\usepackage{amssymb}
\usepackage{hyperref}
\usepackage{graphicx}
\usepackage{lipsum}
\usepackage{lscape}
\usepackage{amsmath}
\usepackage{relsize}%
\usepackage{multirow}
\usepackage[figuresright]{rotating}
\usepackage{lineno}
\usepackage{algorithm} 
\usepackage{algpseudocode} 
\usepackage{comment}
\usepackage{lineno}
\usepackage{nomencl}
\makenomenclature
\usepackage{calc}
\usepackage[T1]{fontenc}
\usepackage{relsize}
\usepackage{subfigure}
\usepackage{tikz}
\usetikzlibrary{fit}
\usepackage{booktabs}
\usepackage{float}

\usepackage{tabularx}
\usepackage{array}

\usepackage{natbib}
\setcitestyle{authoryear,open={(},close={)},citesep={;}} 

\hypersetup{
  colorlinks,
  citecolor=Violet,
  linkcolor=Red,
  urlcolor=Blue}

\newcolumntype{P}[1]{>{\centering\arraybackslash}p{#1}}

\def\tsc#1{\csdef{#1}{\textsc{\lowercase{#1}}\xspace}}
\tsc{WGM}
\tsc{QE}
\tsc{EP}
\tsc{PMS}
\tsc{BEC}
\tsc{DE}

\begin{document}
\let\WriteBookmarks\relax
\def\floatpagepagefraction{1}
\def\textpagefraction{.001}
\shorttitle{Benchmarking YOLO Models for Deer Detection}
\shortauthors{Adhikari et~al.}
 
\title [mode = title]{A Comprehensive Evaluation of YOLO-based Deer Detection Performance on Edge Devices}

\author[1]{Bishal Adhikari}\ead{ba974@msstate.edu}
\author[2]{Jiajia Li}\ead{lijiajia@msu.edu}
\author[3]{Eric S. Michel}\ead{eric.michel@msstate.edu}
\author[3]{Jacob Dykes}\ead{jacob.dykes@msstate.edu}
\author[4]{Te-Ming Paul Tseng}\ead{t.tseng@msstate.edu}
\author[1]{Mary Love Tagert}\ead{mltagert@abe.msstate.edu}
\author[1]{Dong Chen*}\ead{dc2528@msstate.edu}

 \address[1]{Department of Agricultural and Biological Engineering, Mississippi State University, Starkville, MS, USA}
\address[2]{Department of Electrical and Computer Engineering, Michigan State University, East Lansing, MI, USA}
\address[3]{Department of Wildlife, Fisheries and Aquaculture, Mississippi State University, Starkville, MS, USA}
\address[4]{Department of Plant and Soil Science, Mississippi State University, Starkville, MS, USA}
\address{* Dong Chen is the corresponding author}

\begin{abstract}
The escalating economic losses in agriculture due to deer intrusion, estimated to be in the hundreds of millions of dollars annually in the U.S., highlight the inadequacy of traditional mitigation strategies such as hunting, fencing, use of repellents, and scare tactics. This underscores a critical need for intelligent, autonomous solutions capable of real-time deer detection and deterrence. But the progress in this field is impeded by a significant gap in the literature, mainly the lack of a domain-specific, practical dataset and limited study on the viability of deer detection systems on edge devices. To address this gap, this study presents a comprehensive evaluation of state-of-the-art deep learning models for deer detection in challenging real-world scenarios. We introduce a curated, publicly available dataset of 3,095 annotated images with bounding box annotations of deer. Then, we provide an extensive comparative analysis of 12 model variants across four recent YOLO architectures (v8 to v11). Finally, we evaluated their performance on two representative edge computing platforms: the CPU-based Raspberry Pi 5 and the GPU-accelerated NVIDIA Jetson AGX Xavier to assess feasibility for real-world field deployment. Results show that the real-time detection performance is not feasible on Raspberry Pi without hardware-specific model optimization, while NVIDIA Jetson provides greater than 30 frames per second (FPS) with ‘s’ and ‘n’ series models. This study also reveals that smaller, architecturally advanced models such as YOLOv11n, YOLOv8s, and YOLOv9s offer the optimal balance of high accuracy (Average Precision (AP) > 0.85) and computational efficiency (Inference Time < 34 milliseconds).
\end{abstract}
\begin{keywords}
Deer \sep Deer detection \sep Wildlife management \sep Edge devices \sep YOLO object detection \sep NVIDIA Jetson \sep Raspberry Pi \sep YOLOv8 \sep YOLOv9 \sep YOLOv10 \sep YOLOv11
\end{keywords}
\maketitle

\section{Introduction}
\begin{comment}
\textit{This paragraph is about deer, a problem for farmers: 
\- damage costs, references
This paragraph is about traditional approaches, e.g., fencing, xx\- costs, etc.
AI+robotics, wildlife management
\-any attempts
}
\end{comment}

\paragraph{}
% Main Problem
The growing population of the white-tailed deer (\textit{Odocoileus virginianus}, hereafter deer) across the United States is a serious challenge for farmers, as they have to bear significant economic losses from deer depredation of corn, soybeans, cotton, and wheat. For instance, the state of Mississippi has the highest deer density in the nation, where a survey of row crop producers revealed that 17,830 total acres of farmland were affected by deer damage, resulting in a staggering annual economic loss of \$4.6 million \citep{MillsCroft2025}. The issue occurs nationwide. One study estimated that the combined loss due to wildlife damage across corn, soybeans, wheat, and cotton to be \$592.6 million \citep{mckee_estimation_2021}, and another identified white-tailed deer as the primary wildlife species responsible for most of this damage across the U.S. \citep{boyer_unwelcomed_2024}.

% Tradiitonal Solution and their limitations
People have explored a variety of control methods to reduce wildlife-related crop damage, such as fencing, hunting, trapping, and repellents. Each of these approaches, however, presents significant limitations in terms of cost, scalability, and long-term sustainability. Fencing is one of the most widely used deterrents, and high-tensile or electrified fences can be effective in reducing deer damage. However, the costs are prohibitive for large-scale agricultural operations, averaging \$13,000 per mile of fencing, and ongoing maintenance is required to repair damage from storms, fallen trees, or determined animals \citep{Landguth2020}. Moreover, persistent deer often learn to breach or circumvent barriers, reducing effectiveness over time. Hunting and culling programs represent another strategy. Regulated hunting seasons can provide some localized relief, but they are typically seasonal and sometimes of low-intensity, decreasing their effectiveness in reducing deer damage. A recent study on hunting behavior in the U.S. projected that the number of hunters is declining by 2.2\% to 1.5\% per year \citep{Mohr_reducing_hunters}. This further calls into question the long-term efficacy of hunting as a strategy for mitigating deer intrusion on agricultural lands. More intensive damage control programs may reduce local populations more significantly; yet, they often face public opposition, animal welfare concerns, and ecological debates about altering wildlife populations. These programs also require permits from relevant wildlife management agencies, limiting farmers’ autonomy and flexibility \citep{warrenDeer}. 
% Trapping is often employed for wild pigs and can be effective when used strategically with large corral-style traps. Yet this method is highly labor-intensive, requiring constant monitoring, baiting, and maintenance. Trapping also carries the risk of capturing non-target wildlife, raising both ecological and ethical concerns \citep{conover}. In addition, while trapping may temporarily suppress populations in localized areas, wild pigs reproduce rapidly and reinvade treated lands, making trapping alone insufficient as a long-term solution.

Repellents have also been widely studied, particularly for deer \citep{elHani1995}. While chemical or odor-based repellents can reduce browsing pressure, they typically degrade quickly after exposure to rain, wind, or sun and therefore require frequent reapplication. Overall, these approaches are largely \textit{reactive} rather than \textit{proactive}, addressing the problem only after damage has occurred. They are often unsustainable in the long term and may cause unnecessary stress, injury, or mortality to wildlife, raising ethical as well as ecological concerns \citep{dubois2017international}. 

Sustainable wildlife management that actively prevents deer from entering farms represents a foundational solution to these challenges \citep{abed2025iot, currepotential2024}. Future systems can leverage advanced computer vision and deep learning techniques, most notably You Only Look Once (YOLO) models \citep{terven2023comprehensive}, to achieve highly accurate and real-time identification of deer presence and movement \citep{alertsystemsika2024}. Detailed outputs, such as the location of deer within an image or enhanced segmentation masks, can then trigger tailored, dynamic deterrents including species-specific sound or light emissions, or autonomous maneuvers by Unmanned Aerial Vehicles (UAVs) and Unmanned ground Vehicles (UGVs) \citep{roca2024}. 
A common design pattern in emerging research involves a hierarchical process in which initial motion detection, often through Passive Infrared (PIR) sensors, activates more sophisticated vision-based models for animal detection. For example, \cite{GeerthikAnimalDeterrenceIOT} integrated a Convolutional Neural Network (CNN) to identify animals, which then activated non-lethal deterrents such as ultrasonic sounds, flashing lights, or sprinklers. Similarly, \cite{mishra2024smart} employed a PIR sensor to trigger a Recurrent Convolutional Neural Network (R-CNN), which in turn activated species-specific ultrasonic frequencies to mitigate habituation. More advanced systems have also been developed for other species. For instance, \cite{SinghRepellent} designed a wild boar deterrent system that uses YOLOv5n \citep{jocher2020ultralytics} to confirm presence before escalating deterrents from ultrasonic sound to predator scent (wolf urine) and vocalizations. 

Machine vision serves as the foundation for these emerging deterrence systems, acting as the ``eyes'' for real-time deer detection. Among the available approaches, YOLO models have been preferred over two-stage detectors (e.g., Faster R-CNN or EfficientDet), due to their balance of accuracy and speed, making them well-suited for field deployment \citep{i_real-time_2025, li_intelligent_2023}. For instance, \cite{i_real-time_2025} demonstrated the superiority of YOLOv10 over YOLOv5 for wildlife detection on an NVIDIA Jetson Nano, achieving a AP@0.5 of 0.934. Similarly, \cite{v8powered} implemented a YOLOv8m-based surveillance system on a Raspberry Pi, showing its feasibility on low-power edge devices. In a more recent application, \cite{guard2025} deployed a TensorRT-optimized YOLOv5 model on an NVIDIA Jetson Orion Nano for a UAV-based deterrence system, achieving an inference time of only 0.025 seconds per frame. Collectively, these studies affirm that the YOLO family of models has become the de facto standard for practical, high-performance wildlife detection systems on edge hardware.

% The prevalence of YOLO models in wildlife detection and monitoring is well-documented. Researchers have consistently favored these single-stage detectors over more complex architectures like Faster R-CNN or EfficientDet, which often fail to meet the real-time inference demands of edge deployment \citep{i_real-time_2025, li_intelligent_2023}. For instance, \cite{i_real-time_2025} demonstrated the superiority of YOLOv10 over YOLOv5 for wildlife detection on an NVIDIA Jetson Nano, achieving a mAP@0.5 of 0.934. Similarly, \cite{v8powered} implemented a YOLOv8m-based surveillance system on a Raspberry Pi. In a more advanced application, \cite{guard2025} deployed a TensorRT-optimized YOLOv5 model on an NVIDIA Jetson Orion Nano for a UAV-based deterrence system, achieving an inference time of 0.025s per frame. These studies affirm that the YOLO architecture is the de facto standard for building practical, high-performance wildlife detection systems on edge hardware.

% Use of general datasets/ lack of domain-specific public datasets
Despite recent advancements, most deer and wildlife detection studies continue to rely on private datasets or general-purpose object detection datasets. For instance, \cite{abood_revolutionizing_2023} applied YOLOv5 to the PASCAL VOC dataset, while others used aggregated datasets from Roboflow Universe that include multiple animal classes \citep{i_real-time_2025, v8powered}.
In deer-specific research, custom datasets have been used for detecting Sika deer from camera traps with YOLOv8n \citep{alertsystemsika2024} and for identifying various deer species from UAV imagery with YOLOv8-seg \citep{Roca_2025}. However, these datasets are often limited to simplified scenarios focused on targeted deer populations and are generally not publicly available, making reproducibility and broad benchmarking difficult (see Table \ref{tab:deer_detection_papers}). 
% Table~\ref{tab:deer_detection_papers} summarizes representative studies, highlighting the critical need for publicly available, high-quality, domain-specific datasets that reflect the complex and variable conditions of real farms and fields. 

% While the choice of the YOLO architecture is clear, its performance is highly dependent on the training data and the specific application. Many studies leverage general datasets; for example, \cite{abood_revolutionizing_2023} used PASCAL VOC with YOLOv5, while others have used aggregated datasets from Roboflow Universe containing multiple animal classes \citep{i_real-time_2025, v8powered}. Recognizing the limitations of generalized data, other researchers have developed specialized datasets for unique challenges, such as detecting small wildlife targets in UAV imagery \citep{mou_waid_2023}. Within deer-specific research, studies have used custom datasets for detecting Sika deer from camera traps with YOLOv8n \citep{alertsystemsika2024} and various deer species from UAVs with YOLOv8-seg \citep{Roca_2025}. However, these studies typically focus on the performance of a single YOLO model and often utilize private datasets, making direct comparisons difficult.

\begin{table*}[H]
\renewcommand{\arraystretch}{1.5}
\centering
\caption{Dataset and deep learning models across deer detection literature (N/A = Not Accessible).}
\label{tab:deer_detection_papers}
\resizebox{0.99\textwidth}{!}{%
\begin{tabular}{lcclcc}
\toprule
\textbf{Paper} & \textbf{Year} & \textbf{Dataset Size (No. of Images)} & \textbf{Detection Model} & \textbf{Method of Data Acquisition} & \textbf{Data Availability} \\
\midrule
\cite{Roca_2025} & 2025 & 39000 & YOLOv11, RT-DETR & UAV Capture & N/A \\
\midrule
\cite{guard2025} & 2025 & N/A & YOLOv5 & Roboflow & Open Source \\
\midrule
\cite{alertsystemsika2024} & 2024 & 6400 & YOLOv8n & Cameratrap & N/A \\
\midrule
\cite{v8powered} & 2024 & 7000 & YOLOv8n & Roboflow & Open Source \\
\midrule
\cite{yengender} & 2024 & 400 & YOLOv9 & Google Images & N/A \\
\midrule
\cite{roca2024} & 2024 & 39000 & YOLOv8n-seg & UAV Capture & N/A \\
\midrule
\cite{ddlSiddique} & 2023 & 1063 & YOLOv5, ResNet & Mixed & N/A \\
\midrule
\cite{intrusionDetectionImprovement} & 2023 & 4000 & Custom, MobileNet, VGG & N/A & N/A \\
\midrule
\cite{embeddedRepel2021} & 2021 & 1000 & YOLOv3, Tiny-YOLOv3 & Cameratrap & N/A \\
\midrule
\cite{arshadWhere} & 2020 & N/A & YOLOv3 & Manually & N/A \\
\midrule
\cite{munianAuto} & 2020 & 1067 & CNN+HOG Custom & ThermalCam & N/A \\
\bottomrule
\end{tabular}
}
\end{table*}

% Reinforcing gap and connecting to our work
On the other hand, existing studies have primarily focused on individual YOLO models, without providing comprehensive benchmarking across architectures or systematic evaluations. In addition, most of the studies only evaluated the models' performance on high-end GPU devices, and few projects have examined performance on edge devices in terms of efficiency and effectiveness, which is critical for practical field deployment. To address these gaps, we present a publicly available dataset of 3,095 annotated deer images with bounding-box labels, derived from the Idaho Cameratraps project, representing challenging real-world scenarios. Using this dataset, we establish a comprehensive benchmark of YOLOv8–YOLOv11 models, evaluating performance not only on a high-end NVIDIA RTX 5090 GPU but also on resource-constrained platforms, including the CPU-based Raspberry Pi 5 and the GPU-accelerated NVIDIA Jetson AGX Xavier. This research provides a valuable resource for advancing machine vision systems for wildlife detection and control, as well as related applications in agriculture and beyond. The key contributions of this work are as follows: 
\begin{itemize}
    \item An open-source dataset of 3,095 annotated deer images with bounding-box labels, covering diverse environmental conditions and lighting scenarios.  
    \item A comprehensive evaluation and benchmarking of four YOLO architectures (v8–v11), encompassing 12 model variants for deer detection.  
    \item Inference benchmarking of the 12 YOLO model variants on edge devices, including the CPU-based Raspberry Pi 5 and the GPU-accelerated NVIDIA Jetson AGX Xavier.  
\end{itemize}

This paper is organized as follows. Section \ref{sec:methods} describes the data acquisition methods, model selection, training, testing, and inference metrics. Section~\ref{sec:results} presents the model evaluation results, along with the limitations of the study. Finally, Section~\ref{sec:conclusion} provides concluding remarks and potential directions for future work.

\section{Materials and Methods}\label{sec:methods}
This section describes the data acquisition process, the YOLO models employed, the training procedure, and the model evaluation strategy, followed by inference experiments conducted on edge devices.

\begin{figure*}[H]
\centering
{\includegraphics[width=0.95\textwidth]{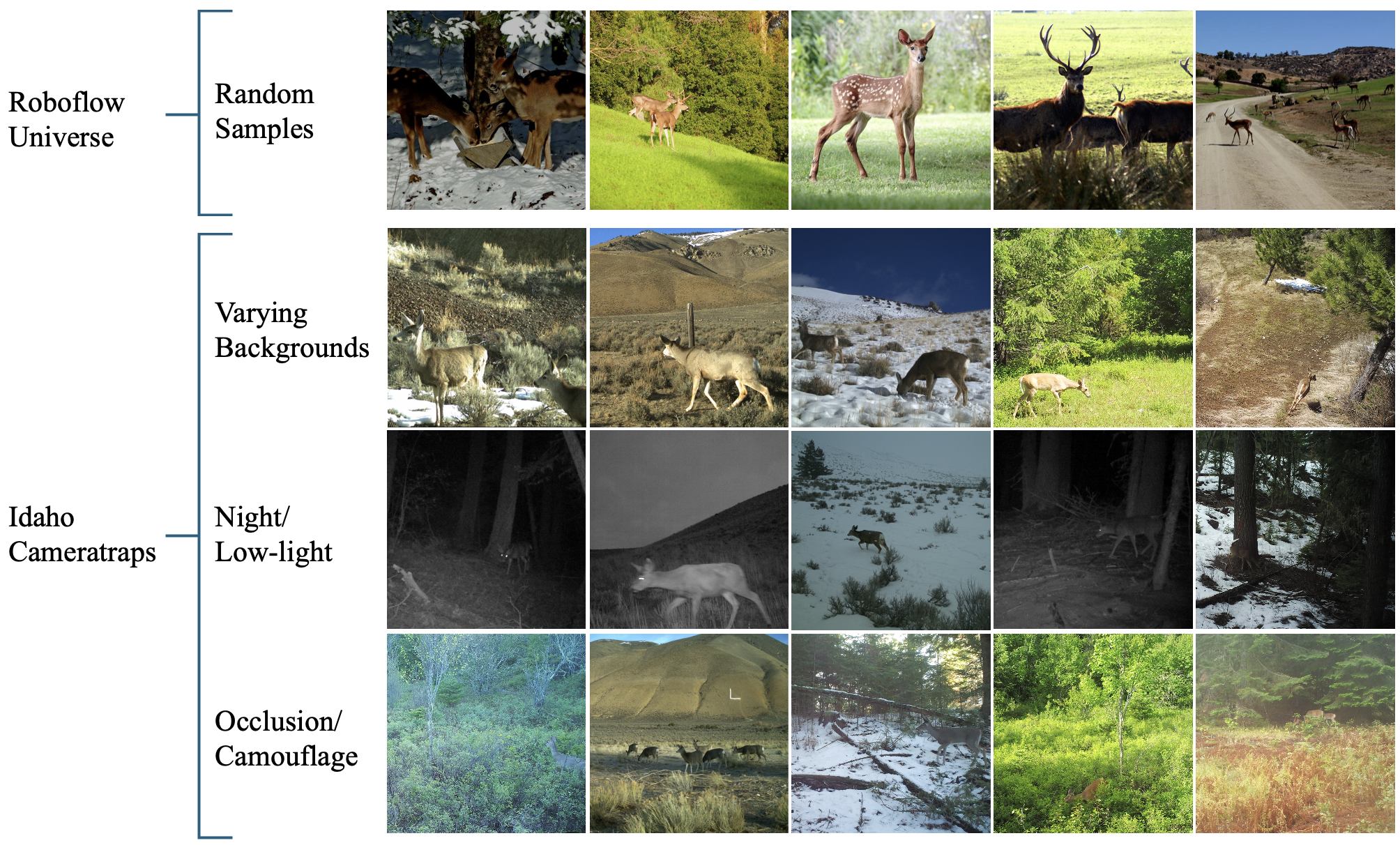} }
    \caption{Sample images from two sources: Roboflow Universe and Idaho Cameratraps.}
    \label{fig:dataset}
    \vspace{-10pt}
\end{figure*}

\subsection{Data acquisition}
To enable a comprehensive comparison, we curated two deer image datasets.  We first selected a dataset from Roboflow Universe containing annotated images of deer, which will be referred to as the ``Roboflow dataset''. The Roboflow dataset \citep{deer-iyhon_dataset} consists of 2,339 images of different deer species captured in diverse environments, mostly under favorable lighting conditions and with minimal motion, as is common in manually photographed images (Figure~\ref{fig:dataset}). The dataset was divided into 2,043 training images and 296 validation images. However, this type of dataset does not fully capture the challenges of real-world field conditions, such as low-light environments, motion blur, partial occlusion, and varying camera trap settings. 

To better represent these scenarios, we curated a second dataset from the Idaho Cameratraps project, shared by the Idaho Department of Fish and Game via the LILA BC online repository \citep{idaho_cameratraps}. The Cameratraps project contains over 1.5 million camera trap images collected from video sequences across multiple regions. While the original dataset provides only sequence-level labels without bounding-box annotations, we filtered approximately 7,000 images containing deer (confidence score > 0.9) and manually annotated 3,095 images with bounding boxes using the Computer Vision Annotation Tool (CVAT) \citep{cvat2024}. This ``Cameratrap dataset'' was then divided into 2,578 training images and 517 validation images (Figure~\ref{fig:dataset}).  

The two datasets differ significantly in complexity and realism. The Roboflow dataset consists of clean, manually photographed images of deer, mostly under good lighting and with limited motion or occlusion, making it more suitable for baseline model training. In contrast, the Cameratrap dataset captures deer under challenging real-world conditions, including varying backgrounds, night/low-light settings, occlusion, and camouflage. As shown in Figure~\ref{fig:dataset}, the Cameratrap dataset therefore provides a more realistic benchmark for evaluating model robustness in field deployments.

\subsection{Deep learning models}
The YOLO framework made a groundbreaking contribution to computer vision, particularly in object detection, by introducing a single-pass approach that framed detection as a regression problem. This innovation enabled real-time inference, which was lacking in other state-of-the-art models at the time \citep{redmon2016lookonceunifiedrealtime}. Since then, the YOLO family has evolved rapidly, producing numerous variants that build on the original design with architectural and performance enhancements \citep{terven2023comprehensive}.

In this study, we focus on four recent versions, YOLOv8, YOLOv9, YOLOv10, and YOLOv11, for comparative benchmarking. The study is further constrained to the three most lightweight variants of each model family: nano (n), small (s), and medium (m). Notably, for YOLOv9, the tiny (t) variant was used because it has a similar parameter count as for nano-scale variants. YOLOv8 introduced notable improvements, including an anchor-free detection method for enhanced accuracy, architectural refinements in the backbone and detection head to better capture objects of varying scales, and a decoupled head design that further improved precision \citep{yaseen2024yolov8indepthexplorationinternal, varghese_yolov8_2024}. We conclude our comparisons with YOLOv11, as subsequent models such as YOLOv12 shift toward attention-centric vision transformer architectures, representing a fundamental departure from the CNN-based architectures of earlier YOLO versions \citep{tian2025yolov12attentioncentricrealtimeobject}.

% You Only Look Once (YOLO)'s groundbreaking contribution in computer vision, specifically in object detection, was its innovative single-pass approach that framed object detection as a regression problem and allowed real-time inference, which was lacking in other top models at that time \cite{redmon2016lookonceunifiedrealtime}. There have been many advancements in the original YOLO model and now we have a sea of models, each different from its predecessor. In this study, we choose to compare the performance of four different versions viz. v8, v9, v10 and v11. YOLOv8 had significant improvements, notably, the anchor-free detection method leading to better overall accuracy than previous models like v5 and v7, improvements in the head and backbone of the model allowing it to capture objects of varying sizes in the scene, and finally a decoupled head improving the accuracy of the model \citep{yaseen2024yolov8indepthexplorationinternal, varghese_yolov8_2024}. We stop at YOLOv11 because newer models like YOLOv12 adopted attention-centric vision transformer architecture, increasing the complexity of the model and are fundamentally different from CNN-based architectures used by its predecessors \citep{tian2025yolov12attentioncentricrealtimeobject}.

\subsubsection{YOLOv8}
YOLOv8, developed by Ultralytics, represents a state-of-the-art object detection model that introduced several architectural improvements over its predecessor, YOLOv5 \citep{hussain2024}. One of its key innovations is anchor-free estimation, which accelerates post-processing, particularly Non-Maximum Suppression (NMS), for selecting prediction boxes \citep{Solawetz_Francesco_2024}. Additionally, the model employs the CSPDarknet backbone, which reduces computational complexity while maintaining accuracy by splitting feature maps and incorporating the Sigmoid Linear Unit (SiLU) activation function \citep{hussain2024}. This design enhances gradient flow and feature reuse, resulting in smaller, more efficient models well-suited for edge devices. Furthermore, YOLOv8 integrates a PANet neck, which improves multi-scale feature learning by augmenting the traditional Feature Pyramid Network (FPN) with a bottom-up path alongside the top-down pathway \citep{reis2024realtimeflyingobjectdetection}. Another improvement is the C2f (Cross Stage Partial network with two fusion blocks) module, which captures complex patterns more effectively, further boosting model accuracy \citep{hussain2024, talibcabv8}. Ultralytics provides YOLOv8 in five sizes (``n'', ``s'', ``m'', ``l'', and ``x''), offering flexibility depending on accuracy and resource constraints. The smallest model, YOLOv8n, has 3.2 million parameters, while the largest, YOLOv8x, has approximately 257.8 million parameters.

\begin{figure*}[H]
\centering
{\includegraphics[width=\columnwidth + \columnwidth]{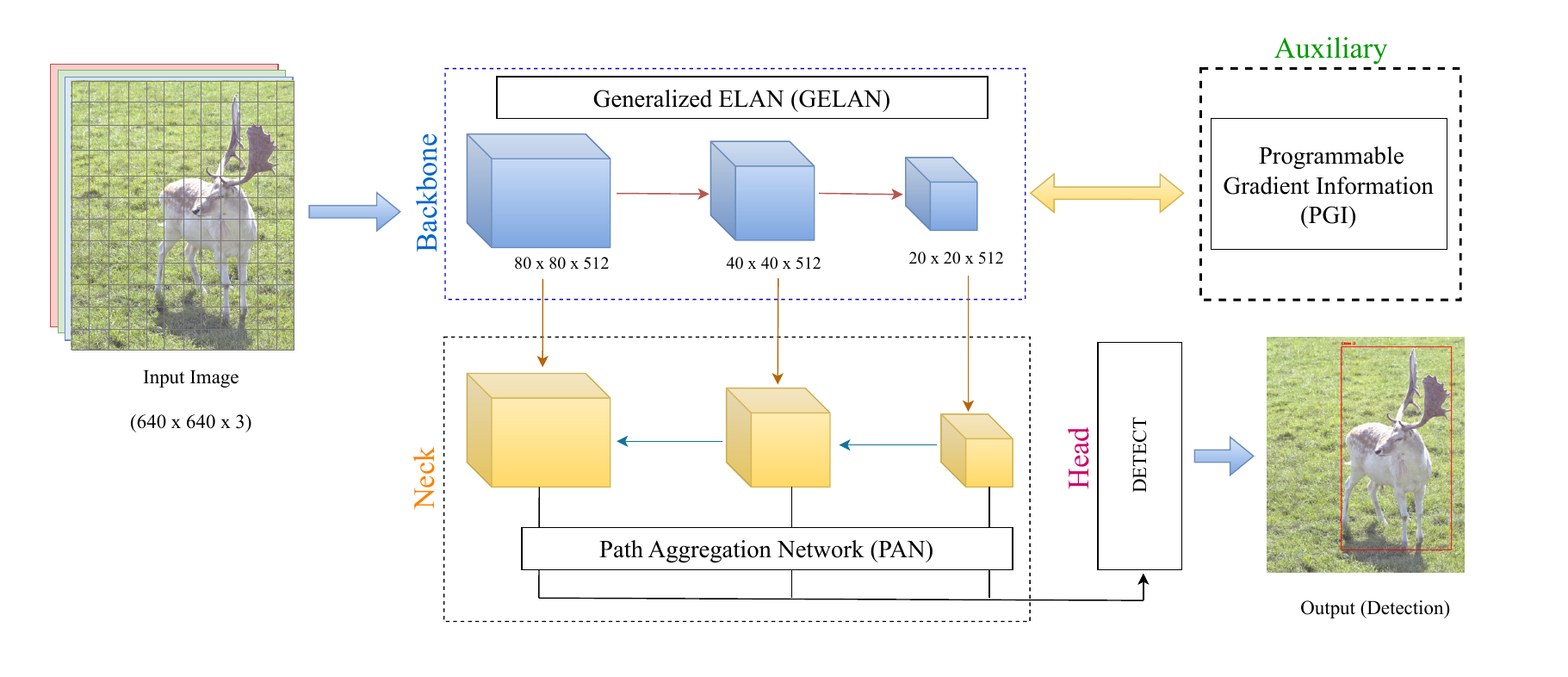} }
    \caption{Overview of YOLOv9 architecture \citep{y9_image}.}
    \label{fig:y9}
    \vspace{-10pt}
\end{figure*}

\subsubsection{YOLOv9} 
YOLOv9 architecture marks a significant improvement over earlier YOLO versions in terms of speed and accuracy of object detection. A key challenge it addressed is the information bottleneck problem in deep neural networks, where gradients diminish or useful information is lost as they propagate through successive layers of large models. To overcome this, YOLOv9 introduced Programmable Gradient Information (PGI), which ensures more reliable gradient flow and enhances the network’s ability to learn from complex image features. As shown in Figure~\ref{fig:y9}, PGI is implemented through a reversible auxiliary branch that updates the main branch by generating useful gradients through a supervision mechanism. Moreover, YOLOv9 introduced the Generalized Efficient Layer Aggregation Network (GELAN), an advancement over the original Efficient Layer Aggregation Network (ELAN) architecture. Unlike ELAN, which relied solely on convolutional blocks, GELAN can flexibly incorporate different computational blocks, improving both efficiency and generalization. Together, the integration of PGI and GELAN enables YOLOv9 to achieve robust gradient flow, efficient computation, and superior detection performance \citep{wang2024yolov9learningwantlearn, parambilv9}.

\subsubsection{YOLOv10}
YOLOv10 introduced several architectural innovations aimed at improving both efficiency and accuracy, while addressing key limitations in earlier YOLO versions \citep{yolov10}. One major issue identified in prior models was the reliance on Non-Maximum Suppression (NMS) during post-processing, which added computational overhead and delayed inference. To overcome this, YOLOv10 adopted NMS-free training by incorporating a one-to-one prediction head alongside the traditional one-to-many head during training, allowing the network to optimize using both. However, at inference time, only the one-to-one head is used, which eliminates the need for NMS and enables faster, end-to-end deployment. Another notable improvement is the simplification of the classification head. The authors observed that the regression head plays a more critical role in YOLO performance, so the classification head was streamlined to reduce computational cost without sacrificing accuracy. 

YOLOv10 also redesigned its convolutional blocks to reduce redundancy. Traditional YOLO architectures used 3×3 convolutions with a stride of 2 to simultaneously downsample spatial dimensions (from H×W to H/2×W/2) and expand channels (from C to 2C). In contrast, YOLOv10 decouples these operations: a pointwise convolution first handles channel transformations, followed by a depthwise convolution for spatial downsampling. This design significantly reduces computational overhead.
To further enhance accuracy, YOLOv10 integrates partial self-attention and employs improved large-kernel convolutions, particularly for lightweight models. Together, these modifications result in a more efficient and accurate object detection framework suitable for real-time applications.

% \cite{yolov10} identified a critical post-processing problem associated with previous YOLO versions, Non-Maximum Suppression (NMS) that utilized time and computation during inference. YOLOv10 introduced significant changes, notably the NMS-free training, lightweight classification head, rank-guided block design, and other efficiency and accuracy-driven improvements. For NMS-free training, YOLOv10 incorporated an one-to-one prediction head alongside the traditional one-to-many head, optimizing the network using both of them during the training. But during inference, only the one-to-one head is used, which makes the model fast and suitable for end-to-end deployment. In YOLOv10, classification head is simplified to reduce its computational overhead since the authors identified that the regression head is more critical for the performance of YOLO models. Traditional convolution blocks in YOLOs perform simultaneous spatial-channel downsampling of image space from $(H \times W)$ to $(H/2 \times W/2)$  and channel space from $C$ to $2C$ using $3\times3$ convolution with stride of 2. In YOLOv10, a decoupled downsampling is achieved by a pointwise convolution followed by a depthwise convolution for spatial and channel transformations respectively. This was found to reduce the computational cost of the operation. With regards to improving the accuracy of YOLOs, YOLOv10 employed partial self-attention and improvised large-kernel convolution for lighter models. 

\begin{table}[H]
\small
\centering
\caption{YOLO model variants used in this study.}
\label{tab:models}
\begin{tabular}{ccc}
\toprule
\textbf{Model Name} & \textbf{Parameters (millions)} & \textbf{GFLOPs} \\
\midrule
YOLOv8n & 3.01 & 8.09 \\
YOLOv8s & 11.13 & 28.44 \\
YOLOv8m & 25.84 & 78.69 \\
\midrule
YOLOv9t & 1.97 & 7.60 \\
YOLOv9s & 7.17 & 26.73 \\
YOLOv9m & 20.01 & 76.51 \\
\midrule
YOLOv10n & 2.27 & 6.53 \\
YOLOv10s & 7.22 & 21.41 \\
YOLOv10m & 15.31 & 58.85 \\
\midrule
YOLOv11n & 2.58 & 6.31 \\
YOLOv11s & 9.41 & 21.30 \\
YOLOv11m & 20.03 & 67.65 \\
\bottomrule
\end{tabular}
\end{table}

\subsubsection{YOLOv11}
YOLOv11 represents the most recent advancement in the YOLO family, developed by Ultralytics as a more sophisticated yet efficient model. It extends the versatility of YOLO by supporting multiple computer vision tasks, including object detection, instance segmentation, and pose estimation, while maintaining strong performance relative to its size \citep{khanam2024yolov11overviewkeyarchitectural}. Built upon the YOLOv8 architecture, YOLOv11 introduces two key innovations. The C2PSA (Cross Stage Partial with Spatial Attention) block enables the model to focus on the most relevant regions of an image, thereby improving its ability to detect small and occluded objects. Meanwhile, the C3k2 (Cross Stage Partial with 3 Convolutions and 2 Kernels) block reduces computational complexity while preserving rich feature representations, making the model more efficient without compromising accuracy.

% YOLOv11 is further advancement to YOLO models and was developed by Ultralytics as a model more sophisticated yet powerful for its size. YOLOv11 is a robust model capable of several computer vision tasks, such as object detection, instance segmentation, pose estimation, etc. YOLOv11 is built upon YOLOv8 architecture and the main improvements in the architecture are C2PSA(Cross Stage Partial with Special Attention) and C3k2(Cross Stage Partial with 3 Convolutions and 2 Kernels) blocks. C2PSA utilizes a spatial-attention mechanism to help the model focus on important regions in the images, improving the performance in detecting small and occluded objects. Similarly, C3k2 is a more efficient version of C3, as it utilizes two kernels to reduce computational complexity while retaining the rich feature representations in an image \cite{khanam2024yolov11overviewkeyarchitectural}. 

Ultralytics maintains the YOLO family of models and provides open-source implementations of the versions they primarily developed, such as YOLOv8 and YOLOv11. In practice, YOLO models present a tradeoff between model size and reliability. Although they are widely applicable across various computer vision tasks, our study focuses specifically on their object detection capabilities. To capture the balance between performance and computational efficiency, we selected the ``n'', ``s'', and ``m'' variants of each model. Among these, YOLOv8m is the largest, with 25.84 million parameters, while YOLOv9t is the smallest, with only 1.94 million parameters (see Table~\ref{tab:models}). Each version incorporates different techniques for gradient flow, optimization, and image processing. By testing performance across a wide range of architectural variations and model sizes, we will provide deeper insights that support informed decision-making when selecting YOLO models for deer detection tasks.

\subsection{Training} \label{sec:sub:training}
The YOLO models were trained on an NVIDIA GeForce RTX 5090 GPU with 32 GB of VRAM, a high-end platform well-suited for deep learning workloads. The training was conducted on a Linux machine with CUDA version 12.9 (Table ~\ref{tab:computer_specs}). While the default training configurations provided by Ultralytics are effective, adjustments were made to better leverage the available hardware. In particular, the batch size and number of workers were tuned for different models based on their parameter size. Larger batch sizes enable faster training by improving GPU utilization \citep{ijcai_training}, but they are constrained by memory usage, especially for larger models. Therefore, careful experimentation was conducted to identify the optimal values for batch size and the number of workers, given the model size and available VRAM. All models were trained for a total of 100 epochs under these optimized settings.

\begin{table}[H]
\small
\centering
\caption{Specifications of the device used for model training.}
\label{tab:computer_specs}
% Set the table width to the column width and make the last column wrap.
\begin{tabularx}{0.95\columnwidth}{ll >{\raggedright\arraybackslash}X}
\toprule
\textbf{Item} & \textbf{Specification} & \textbf{Value} \\
\midrule
\multirow{5}{*}{CPU} & Model & AMD Ryzen Threadripper PRO 7965WX \\
& Cores & 24 \\
& Threads & 48 processors (2 threads per core) \\
& Architecture & x86\_64 \\
& Boost Clock & 5.3 GHz \\
\midrule
\multirow{4}{*}{GPU} & Model & NVIDIA GeForce RTX 5090 \\
& VRAM & 32 GB \\
& Power Limit & 600 W \\
& CUDA & 12.9 \\
\midrule
\multirow{2}{*}{Memory} & Total RAM & 125 GB \\
& Available RAM & 115 GB \\
\midrule
System & Platform & Linux \\
\bottomrule
\end{tabularx}
\end{table}

% No new package needed for this basic version

\subsection{Testing and performance evaluation}\label{sec:metrics}
In this study, the model needs to correctly identify the presence of deer in an image and localize them with bounding boxes. In addition, we need to assess the computational efficiency of the models and the feasibility of real-time deployment. There are several standard metrics to evaluate the performance of models.

\textit{Intersection over Union (IoU):} IoU is a widely adopted metric used to evaluate the correctness of detections produced by an object detection model. The model is trained to predict bounding boxes around objects in an image, and IoU measures how closely these predictions match the ground-truth bounding boxes \citep{padilla2020survey}.

\textit{Precision (P)}: Precision quantifies the proportion of detections that are correct, i.e., true positives relative to all predicted positives. In object detection, a detection is considered a true positive if both the class label and localization ($IoU \geq \text{threshold}$) are correct.

% \textit{Precision (P):} Precision measures the ability of the model to correctly identify detected objects as a particular class. It is always calculated for a particular class. It is calculated using the formula:

% \begin{equation}
%     Precision(P) = \frac{TP}{TP + FP}
% \end{equation}
% where $TP=$ True Positives, $FP=$ False Positives.

% In object detection, true positives does not simply mean whether the model classified an object in the image to a correct class. Instead, it also incorporates the correctness of the localization of that object within the image. As discussed above, this is given by IoU metric.
% \begin{itemize}
%     \item Thus, $TP$ measures the total number of detections classified correctly as the class and $IoU \geq threshold$.

%     \item Similarly, $FP$ measures the total number of detections which were incorrectly classified as this class or if $IoU < threshold$. 
% \end{itemize}

\textit{Recall (R)}: Recall measures the proportion of ground-truth objects correctly detected, i.e., true positives relative to all actual objects. Missed detections or those failing the \textit{IoU} threshold contribute to false negatives.

% \textit{Recall (R) : } Recall measures the ability of the model to detect all objects of a particular class in images. It is calculated as :
% \begin{equation}
%     Recall (R) = \frac{TP}{TP + FN}
% \end{equation}
% where $TP=$ True Positives, $FN=$ False Negatives

% While $TP$ here is the same as previously discussed, the new term $FN$ measures the number of objects of a particular class that the model either classified wrongly or for which localization was not good enough, i.e, $IoU < threshold$. Since we have only one class i.e. Deer in our case, $FN$ can include cases where the model completely missed to detect the deer or those cases where the detections did not qualify as a proper detection due to IoU threshold.

\textit{F1 Score}: The F1-score is the harmonic mean of precision and recall, providing a single balanced measure of detection accuracy at a given confidence threshold.

\textit{AP (Average Precision)}: AP summarizes the tradeoff between precision and recall across confidence thresholds by computing the area under the precision–recall curve. Higher AP values indicate better overall detection performance \citep{padilla2020survey}.

% F1 score combines both precision and recall to give a more robust value for the model accuracy. It is calculated as:
% \begin{equation}
%     F1 = 2 \times \frac{Precision \times Recall}{Precision + Recall}
% \end{equation}

% Generally, F1 score is calculated at a confidence threshold. A confidence threshold is the value that decides whether a detection belongs to a certain class based on the confidence score predicted by the model. 

% \textit{AP (Average Precision)}
% Precision, recall and f1 score are calculated for a model with a fixed confidence threshold. But, these values can change if the confidence threshold of the model is changed. To better evaluate the model, measuring precision and recall at different confidence thresholds and plotting them on precision vs recall graph is crucial. A higher confidence threshold generally means that a model will be more cautious before making a positive prediction, consequently increasing the precision but reducing the recall. Precision-recall curve makes this nuance clearer by allowing to visualize the tradeoff between precision and recall.

% Average Precision (AP) metric is the area under the precision-recall curve which gives a single number that summarizes this tradeoff across confidence thresholds. Here, higher $AP$ signifies higher values for both precision and recall indicating a robust model.

\textit{Mean Average Precision (mAP)}: mAP is the mean of AP across all predicted classes. In this study, it reduces to the AP of the single ``Deer'' class. Commonly reported values include mAP@0.5 (IoU threshold of 0.5) and mAP@50\_95 (averaged across thresholds from 0.5 to 0.95 with step size 0.05).

% \textit{Mean Average Precision (mAP)}
% mAP is simply the average of AP across all classes that the model can predict. For our case, this is simply the AP of the class 'Deer'. It is important to note that, mAP is calculated at a certain IoU threshold. Most common is mAP@0.5 which means mAP of the model at IoU threshold of 0.5. Moreover, metrics like mAP@.5:.95 take the mean of AP at IoU threshold values between 0.5 to 0.95 (step size of 0.05).

\textit{Inference Time}: Inference time is the average time required for a model to process an input image and generate detections. It reflects computational efficiency and determines the feasibility of real-time deployment.

% \textit{Inference Time}
% To understand the deployability of a deep learning model, it is crucial to understand its inference speed. Inference speed is the time it takes for a model to pass an image through to generate an output. For YOLO object detection model, when an image is passed through the model, it takes time for the model to perform matrix multiplication on the image array to produce a result containing bounding boxes with confidence scores. Inference time is the sole time consumed by a model to compute upon the input image. In time-critical deployments like live object detections, inference time is significant to the application.

\textit{Processing Time}: Processing time includes additional steps outside inference, such as image pre-processing, non-maximum suppression, and visualization of bounding boxes. Together with inference time, it determines the total detection time. Combined with inference time, it defines the total detection time, expressed as: $\text{Total Time} = \text{Inference time} + \text{Processing Time}$.

\textit{Frames Per Second (FPS)}: FPS is derived from total detection time and indicates the maximum frame rate achievable in live video applications on a given device.

% \textit{Processing Time }

% An image needs to be processed before feeding it to the detection model. The output of the model needs to be processed as well to refine the detections. Since most of the YOLO models predict multiple boxes for a single object in the image, non-maximum suppression is applied to get the single prediction. Even this needs to be processed to create human-understandable forms such as drawing predicted bounding boxes over the original image. All these auxiliary processes to the inference which become useful in the application or deployment, takes time and we refer to this time as processing time in this study. Total time taken for each model to make predictions is thus calculated as below.

% \begin{equation}
%     Total \ Time = Inference \ time \ + Processing \ Time 
% \end{equation}

% \begin{equation}
%     FPS \ = \ \frac{1000}{Total \ Time (ms)}
% \end{equation}
% Total detection time helps us calculate the maximum fps(frames per second) that we can achieve on a live video. It is important to understand that this overall speed is subject to the computing device.

In this study, we use \textit{IoU}, precision, recall, F1-score, and AP@0.5 to measure detection accuracy, while inference time and processing time are used to evaluate computational speed \cite{padilla2020survey}.

\subsection{Inference on edge devices}
The YOLO models were trained on high-performance workstations. However, for applications such as deer detection, real-time identification and localization are essential for system effectiveness. One approach is to outsource inference to a cloud-based server, where live video frames are continuously transmitted and processed, with detection results returned to the client \citep{bandaru_review_2024}. However, such solutions often face challenges related to bandwidth requirements, latency, and network security.  

\begin{figure}
    \centering
    \includegraphics[width=0.95\columnwidth]{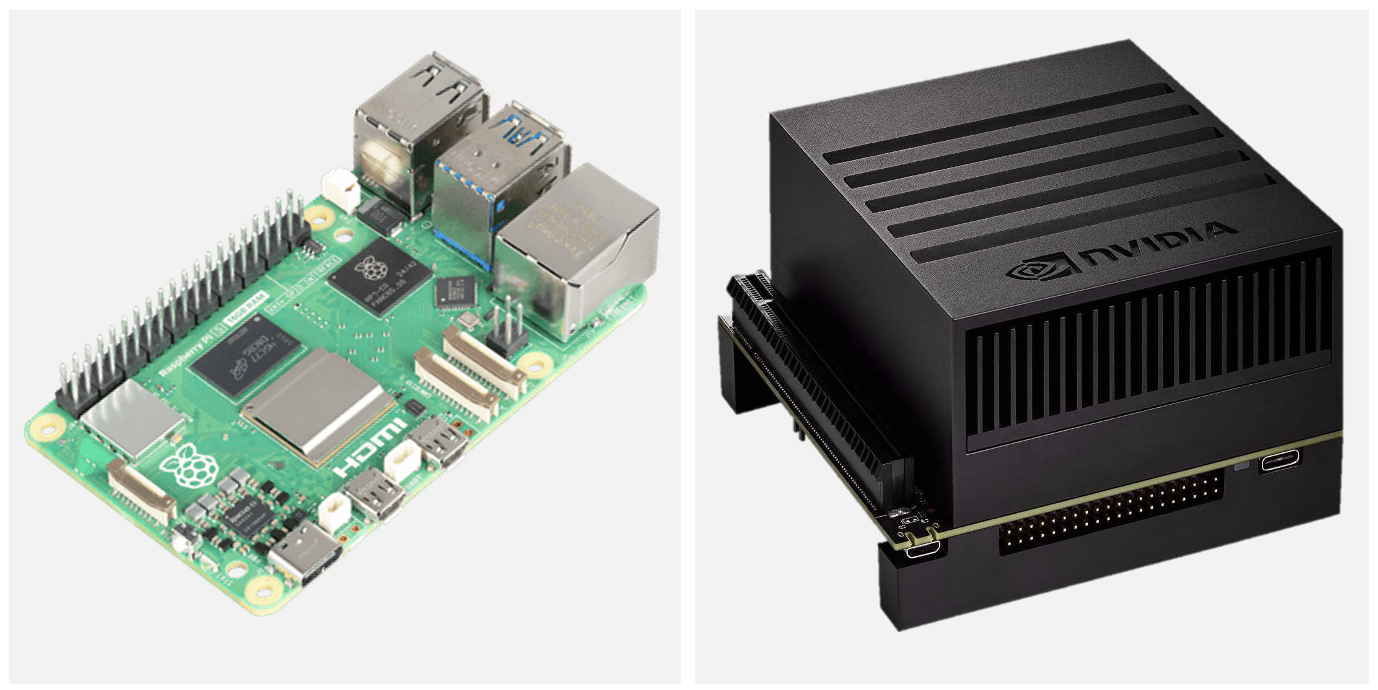}
    \caption{ Raspberry Pi 5 (left) and NVIDIA Jetson AGX Xavier (right).}
    \label{fig:devices}
\end{figure}

Edge computing provides an alternative by enabling local image or video processing, thereby reducing latency and ensuring rapid response. Performing all computation on the local device also creates an independent detection and tracking system, which is particularly advantageous in remote agricultural settings where reliable cloud connectivity cannot be guaranteed \citep{off-the-shelf}. A wide range of off-the-shelf edge devices are available for such applications, including single-board computers (SBCs) such as the Raspberry Pi, NVIDIA Jetson platforms, USB accelerators, and mobile devices with embedded CPUs and GPUs \citep{alqahtani2024benchmarkingdeeplearningmodels}. We evaluate inference performance on two representative devices: a CPU-based Raspberry Pi and a GPU-accelerated NVIDIA Jetson platform (Figure~\ref{fig:devices}).   

% In applications such as deer detection, real-time deer identification and localization are necessary for the overall system to be effective. One way to manage the system is to outsource the inference to a cloud-based server and continuously communicate live video frames with it and get the detection results. But, such solutions often fall short due to challenges in meeting bandwidth requirements, low latency, and network security aspects. On the other hand, edge computing enables local image or video processing and facilitates rapid response. Having all the computing on the local device creates an independent tracking system which is often desired in remote agricultural settings where connecting to cloud services reliably is often challenging \cite{off-the-shelf}. A variety of off-the-shelf edge devices including single board computers (SBCs) like Raspberry Pi, NVIDIA Jetson platforms, as well as USB accelerators, and mobile phones with embedded CPU and GPU are available for these applications \cite{alqahtani2024benchmarkingdeeplearningmodels}. For this study, we perform inference evaluation on a Raspberry Pi which uses CPU and a NVIDIA Jetson Platform with GPU. The details of the devices are discussed below and the details of inference and evaluation criteria are discussed in Section \ref{sec:results}.

\subsubsection{Raspberry Pi 5}
Raspberry Pi is a series of small single-board computers that offer low-cost, small-sized hardware for computing. They are often credit-card-sized but deliver excellent computing power for their size. In this study, we test the performance of trained models on Raspberry Pi 5, which is the newest model of the Raspberry Pi series (Figure \ref{fig:devices}). It features an onboard computer with a quad-core 64-bit Arm Cortex-A76 CPU. Raspberry Pi is widely used in applications like real-time image or video processing since it supports a dedicated Camera Serial Interface (CSI) . Moreover, it can operate on low power, making it suitable for remote deployment as an independent computer, able to get real-time data from the environment, such as the presence of deer in our case. For this study, a device with 16 GB system memory and the operating system Linux Ubuntu 22.04 was used. 
% YOLO models are especially optimized for real-time applications, but the Arm Cortex-A76 CPU, while capable, cannot match the performance of desktop-grade processors, resulting in significant latency in the inference of large, unoptimized models. Similarly, sustained high-intensity workloads can cause the CPU to automatically reduce its clock speed to reduce thermal damage in a process known as thermal throttling. These limitations exist in the Raspberry Pi, which is what necessitates the evaluation of the models so that the most efficient yet highly performant model can be selected for specific deployment purposes.

\subsubsection{Jetson AGX Xavier}
NVIDIA Jetson AGX Xavier is a powerful edge device designed explicitly to address the rigorous demands of robotics and edge AI applications. It is centered around a highly integrated System-on-Chip (SoC) that combines Volta GPU, ARM-based CPUs, and specialized hardware accelerators within a unified memory framework. The GPU incorporates 512 CUDA cores and 64 tensor cores, capable of delivering up to 11 TFLOPS (FP16) or 22 TOPS (INT8). It operates at a maximum frequency of 1.37 GHz with dynamic scaling, enabling fine-grained control of power and performance \citep{cetre:hal-03693764}.  The tensor cores provide significant efficiency for AI workloads, as they are optimized for matrix multiply–accumulate (MMA) operations, which underpin most deep learning algorithms. Moreover, the AGX Xavier integrates dedicated accelerators such as Deep Learning Accelerators (DLA) and Programmable Vision Accelerators (PVA), providing hardware-level support for diverse workloads in autonomous and embedded systems. In addition, this device comes with a compact form factor of 4.2 x 4.2 x 4 inches (Figure \ref{fig:devices} (right)). Together, these features make the Jetson AGX Xavier a versatile platform for real-time robotics and edge AI deployment.

\subsubsection{Model deployment and optimization}
While powerful hardware provides the foundation for high-speed computing, the performance of AI models is equally dependent on their software implementations \citep{ravi2025improvingreproducibilitydeeplearning}. To fully exploit the capabilities of the edge devices, specialized software frameworks and optimizations are required. Ultralytics provides YOLO models with optimizations for accuracy and speed; however, the default PyTorch \citep{paszke2019pytorch} exports are not always the most efficient for deployment on storage- and performance-constrained devices. To address this, several open-source model optimization methods enable faster inference without compromising accuracy. Common export and deployment formats include ONNX, OpenVINO, TorchScript, TensorFlow Lite, NCNN, and TensorRT \citep{iqbal2024review}. Each format supports different types of AI models while incorporating optimizations to improve inference speed on specific hardware. In many cases, the creators of these model frameworks also provide dedicated runtime platforms to ensure efficient execution across devices.

% Even though a highly capable hardware is the foundation for high speed computing, the performance of AI models on these devices is equally influenced by the software implementations \citep{ravi2025improvingreproducibilitydeeplearning}. It is crucial to exploit the capabilities of this hardware through special software programs to be able to get the most out of the device . For YOLO model deployment, there are various frameworks and platforms, some of which are general while some are specialized towards edge devices and specific hardware. Ultralytics generally provides the models with optimizations for accuracy and speed, but the default PyTorch model exported from Ultralytics might not be the most efficient way of deploying these models, especially in storage and performance-constrained edge devices. There exist several open source model optimization methods that can produce significantly faster models without compromising the accuracy and precision of the models. Common model export formats include ONNX, OpenVINO, TorchScript, TF Lite, NCNN, TensorRT, etc. Each of these model formats can represent a variety of AI models but with several optimizations leading to a better inference speed on different hardware devices. Often, the model framework designers also develop the software platforms to efficiently run those models on different hardware devices.

In this study, we evaluated the YOLO models on both devices by converting them to ONNX format. ONNX provides a framework-agnostic representation of deep learning models as computation graphs, enabling portability across platforms \citep{joshua_onnx}. Inference was performed using ONNX Runtime, a lightweight, high-performance engine that reduces deployment overhead and supports hardware-specific optimizations via execution providers (e.g., ARM CPUs or CUDA GPUs). These features make ONNX a widely adopted choice for efficient model deployment.

\begin{table*}[!ht]
\small
\centering
\caption{Performance of models trained on the Roboflow dataset and tested on the Cameratraps dataset. }
\label{tab:model_data_compare}
\begin{tabular}{lcccccc}
\toprule
\textbf{Model} & \textbf{Test Dataset} & \textbf{Precision} & \textbf{Recall} & \textbf{AP@0.5} & \textbf{AP@50\_95} \\
\midrule
YOLOv10s & \multirow{3}{*}{RoboFlow} & 0.9868 & 0.9594 & 0.9910 & 0.8484 \\
YOLOv11n & & 0.9894 & 0.9797 & \underline{\textcolor{ForestGreen}{0.9920}} & \underline{\textcolor{ForestGreen}{0.8475}} \\
YOLOv8n & & 0.9867 & 0.9847 & 0.9918 & 0.8365 \\
\midrule
YOLOv10s & \multirow{3}{*}{Cameratraps} & 0.9032 & 0.5672 & 0.7334 & 0.5356 \\
YOLOv11n & & 0.9306 & 0.5934 & \underline{\textcolor{red}{0.7395}} & \underline{\textcolor{red}{0.5341}} \\
YOLOv8n & & 0.9227 & 0.5773 & 0.7199 & 0.5035 \\
\bottomrule
\end{tabular}
\end{table*}

\section{Experiments and Results}\label{sec:results}
% In this section, we first evaluate model performance when trained on the Roboflow dataset and tested on the Cameratraps dataset, highlighting the differences between an existing dataset and our curated dataset. We then present results obtained on a high-end GPU workstation, followed by inference performance on two representative edge devices.
\begin{figure*} \label{fig:example_detections}
    \centering
{\includegraphics[width=.87\linewidth]{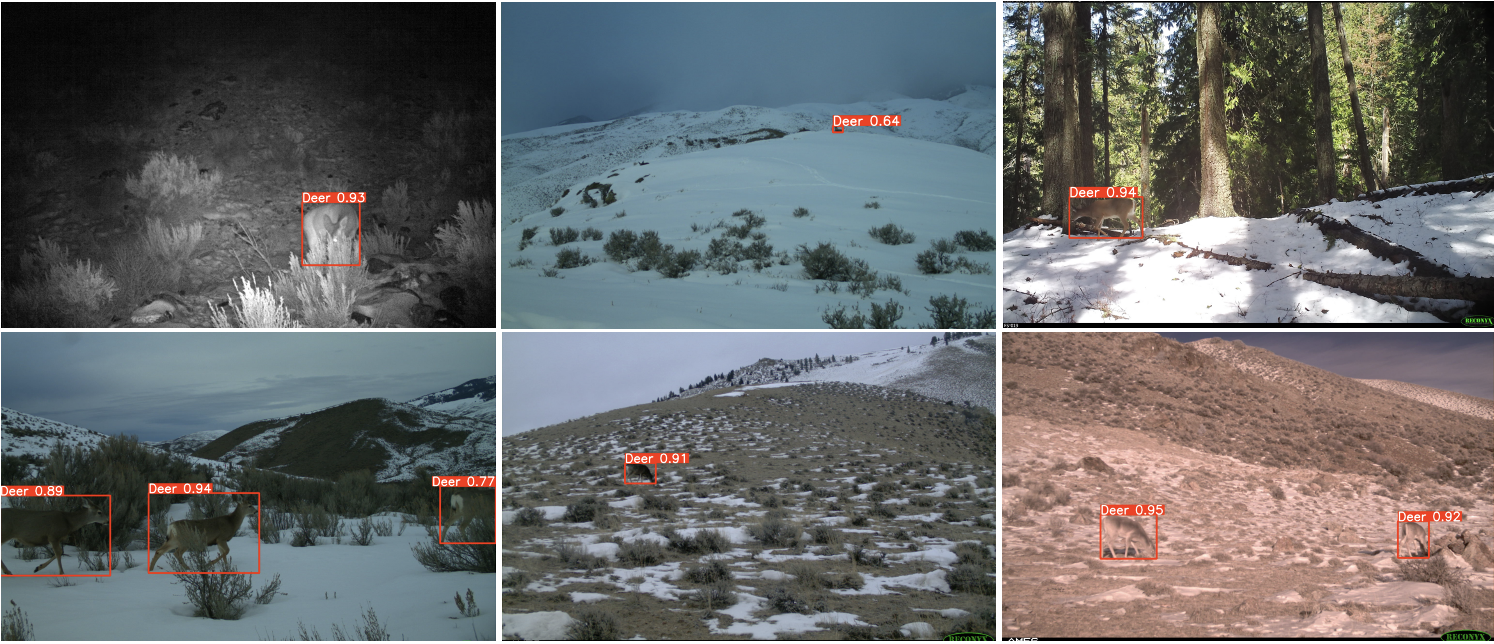} }
    \caption{Example deer detections using YOLOv11m on Cameratraps dataset.}
    \label{fig:example_detect}
    \vspace{-10pt}
\end{figure*}

\subsection{Cross-dataset evaluation}
To evaluate the contribution of our curated dataset, we performed a cross-dataset experiment on the high-end GPU (Table~\ref{tab:computer_specs}) by training models on the Roboflow dataset and testing them on the Cameratraps dataset. As shown in Table~\ref{tab:model_data_compare}, all models achieved excellent performance when trained and tested on the Roboflow dataset, with YOLOv11n reaching AP@0.5 of 0.9920 and AP@50\_95 of 0.8475. However, when the same models were tested on the Cameratraps dataset, performance dropped sharply. For instance, YOLOv11n achieved only 0.7395 AP@0.5 and 0.5341 AP@50–95, despite its near-perfect results on Roboflow. This substantial performance gap highlights the limitations of existing datasets such as Roboflow, which consist mostly of clean, well-lit images, and do not capture the challenging conditions present in real-world scenarios. In contrast, the Cameratraps dataset includes varied lighting, occlusion, camouflage, and motion blur, making it a more realistic benchmark for wildlife detection. Figure~\ref{fig:example_detect} shows some successful detections by the YOLOv11m model on the Cameratraps testing dataset across different scenarios. These images are carefully curated difficult scenarios, but the model shows an impressive ability to handle varied lighting conditions, distances, and deer poses.
Therefore, this evaluation confirms the necessity of using domain-representative datasets. For the remainder of this study, we adopt the Cameratraps dataset as the primary evaluation dataset to ensure that results reflect real-field applicability.

\begin{table*}[!ht]
\small
\centering
\caption{Performance and complexity of YOLO models on different edge devices.}
\label{tab:model_device_compare_final}
\begin{tabular}{l ccc cc ccc cc}
\toprule
\multirow{2}{*}{\textbf{Model}} & \multicolumn{3}{c}{\textbf{Inference (ms)}} & \multicolumn{2}{c}{\textbf{FPS}} & \multicolumn{3}{c}{\textbf{AP@0.5}} & \multirow{2}{*}{\textbf{GFLOPs}} & \multirow{2}{*}{\textbf{Params (M)}} \\

\cmidrule(lr){2-4} \cmidrule(lr){5-6} \cmidrule(lr){7-9}
& \textbf{Jetson} & \textbf{Pi 5} & \textbf{High-end GPU} & \textbf{Jetson} & \textbf{Pi 5} & \textbf{Jetson} & \textbf{Pi 5} & \textbf{High-end GPU} & \\

\midrule

y11n & 15 & 219 & 1.4 & 68 & \underline{\textcolor{ForestGreen}{5}} & 0.8677 & 0.8351 & 0.8439 & \underline{\textcolor{ForestGreen}{6.31}} & 2.58 \\

y10n & \underline{\textcolor{ForestGreen}{13}} & 247 & \underline{\textcolor{ForestGreen}{1.2}} & 75 & 4 & \underline{\textcolor{red}{0.8056}} & \underline{\textcolor{red}{0.8149}} & \underline{\textcolor{red}{0.8056}} & 6.53 & 2.27 \\

y9t & 18 & 267 & 1.8 & 54 & 4 & 0.8367 & 0.8447 & 0.8368 & 7.60 & \underline{\textcolor{ForestGreen}{1.97}} \\

y8n & \underline{\textcolor{ForestGreen}{13}} & \underline{\textcolor{ForestGreen}{217}} & 1.4 & \underline{\textcolor{ForestGreen}{76}} & \underline{\textcolor{ForestGreen}{5}} & 0.8397 & 0.8711 & 0.8411 & 8.09 & 3.01 \\

y11s & 28 & 515 & 1.7 & 36 & 2 & 0.8581 & 0.8869 & 0.8581 & 21.30 & 9.41 \\
y10s & 25 & 472 & 1.4 & 41 & 2 & 0.8588 & 0.8933 & 0.8601 & 21.41 & 7.22 \\
y9s & 30 & 544 & 2.0 & 34 & 2 & 0.8657 & 0.8710 & 0.8657 & 26.73 & 7.17 \\
y8s & 23 & 494 & 1.6 & 43 & 2 & 0.8699 & 0.8670 & 0.8699 & 28.44 & 11.13 \\
y10m & 54 & 985 & 1.8 & 18 & \underline{\textcolor{red}{1}} & 0.8512 & 0.8597 & 0.8512 & 58.85 & 15.31 \\
y11m & 65 & \underline{\textcolor{red}{1702}} & 2.4 & 15 & - & 0.8677 & 0.8755 & 0.8676 & 67.65 & 20.03 \\
y9m & \underline{\textcolor{red}{71}} & 1120 & \underline{\textcolor{red}{2.6}} & \underline{\textcolor{red}{14}} & - & \underline{\textcolor{ForestGreen}{0.8760}} & \underline{\textcolor{ForestGreen}{0.9008}} & \underline{\textcolor{ForestGreen}{0.8760}} & 76.51 & 20.01 \\
y8m & 55 & 1075 & 2.2 & 18 & - & 0.8607 & 0.8917 & 0.8620 & \underline{\textcolor{red}{78.69}} & \underline{\textcolor{red}{25.84}} \\

\bottomrule

\end{tabular}
\end{table*}

\subsection{Raspberry Pi 5 (CPU-based inference)}
Table~\ref{tab:model_device_compare_final} summarizes the performance of the YOLO models on the Raspberry Pi 5. With only CPU support, inference times were orders of magnitude slower than on the benchmark GPU. Even the smallest model, YOLOv9t, required over 250 ms per inference, corresponding to fewer than 4 frames per second (FPS). Larger models, such as YOLOv8m, exceeded 1.5 seconds per frame. These low frame rates are inadequate for monitoring or tracking moving animals, making CPU-only deployment impractical for real-time applications on this class of hardware. 

\begin{figure}
    \centering
    \includegraphics[width=0.95\columnwidth]{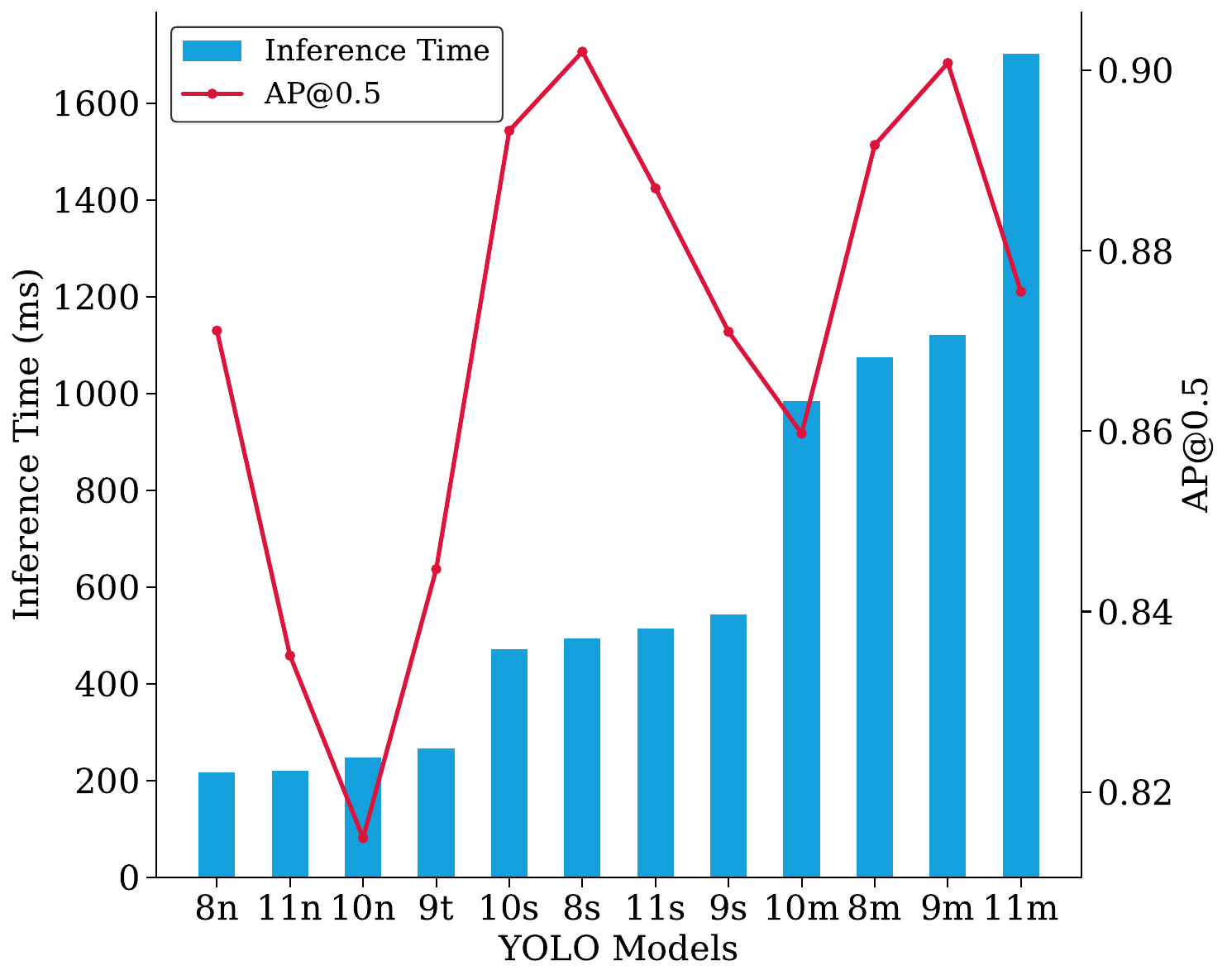}
    \caption{Latency-reliability trade-off across models on Raspberry Pi.}
    \label{fig:reliability_pi}
\end{figure}

Figure~\ref{fig:reliability_pi} shows the latency–reliability trade-off for YOLO models on the Raspberry Pi 5. The bar plot represents inference time (\textit{ms}), while the line plot indicates AP@0.5. The results illustrate the inherent trade-off between detection accuracy and computational latency: smaller models, such as YOLOv9t, achieve lower latency but at the cost of reduced accuracy, whereas larger models like YOLOv8m and YOLOv11m provide higher accuracy but suffer from prohibitive inference times. This also highlights the challenge of balancing model complexity and deployability on CPU-only edge devices.

The performance results, detailed in Table \ref{tab:model_device_compare_final}, highlight the significant computational challenge this presents. While the detection accuracy (AP@0.5) remains unchanged, as it is a function of the model's weights, the inference times are orders of magnitude higher than on the benchmark GPU as evident from the table. Even the smallest model, YOLOv9t, required over 250 ms for a single inference, resulting in a detection speed of less than 4 frames per second (FPS). For the larger models, such as YOLOv8m, the total time exceeded 1.5 seconds per frame. Such low frame rates are insufficient for monitoring or tracking moving animals, rendering most models impractical for any application requiring real-time responsiveness on this class of hardware.

\subsection{NVIDIA Jetson AGX Xavier (GPU-Accelerated Inference)}
\begin{figure}
    \centering
    \includegraphics[width=0.95\columnwidth]{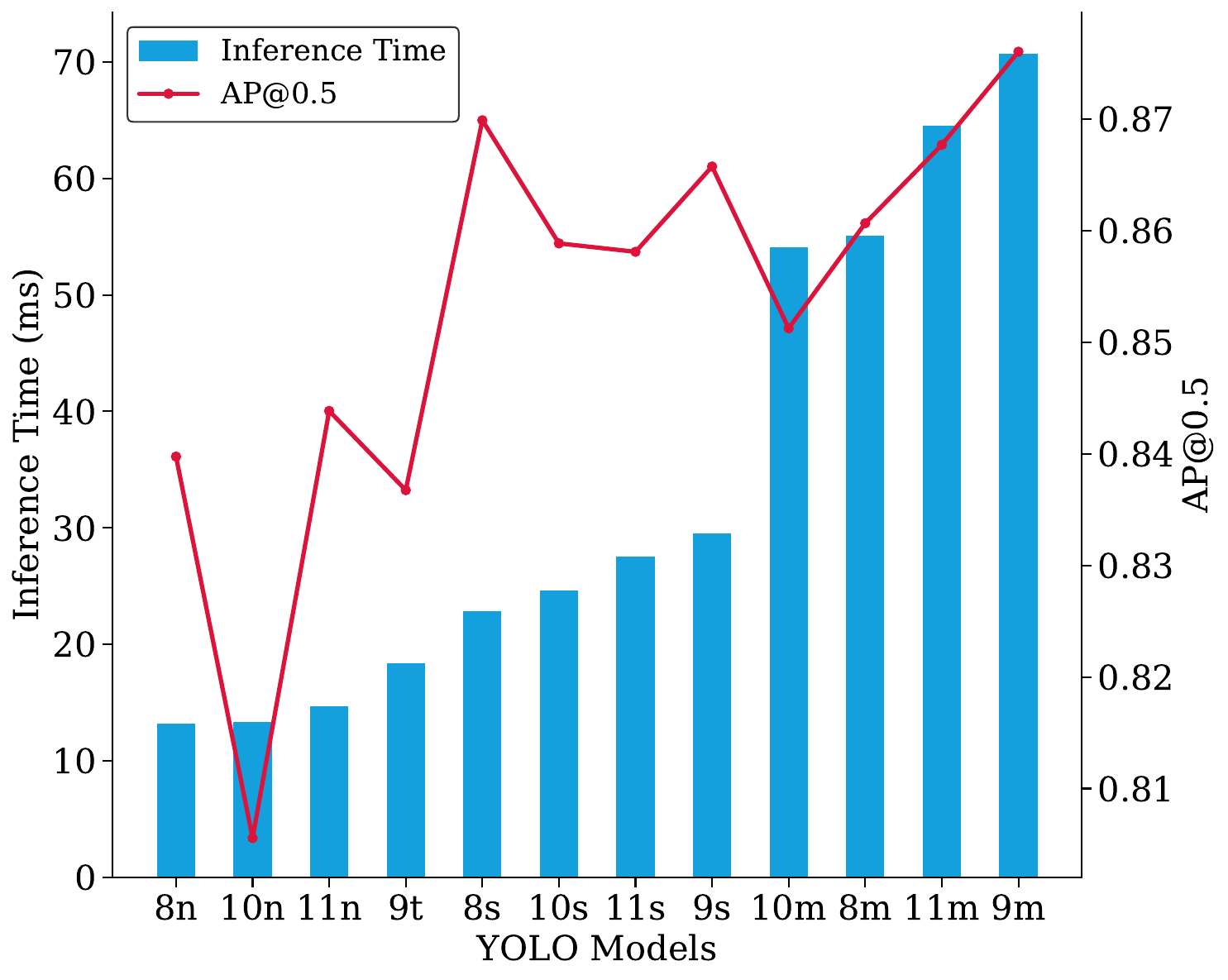}
    \caption{Latency-reliability trade-off across models on NVIDIA Jetson AGX Xavier.}
    \label{fig:reliability_jetson}
\end{figure}

CPU-based edge devices such as the Raspberry Pi 5 face severe latency constraints. In contrast, as shown in Table~\ref{tab:model_device_compare_final}, the Jetson AGX Xavier achieved substantially lower inference times across all models, demonstrating the advantage of the Volta GPU architecture. The smallest models, such as YOLOv11n and YOLOv10n, delivered detection speeds exceeding 40 FPS, comfortably meeting the requirements for real-time video analysis. Even medium-sized models, including YOLOv11s and YOLOv8s, sustained performance above 20 FPS. 
Figure~\ref{fig:reliability_jetson} further illustrates the latency–reliability trade-off across models on the Jetson platform. Inference times remained below 70 ms even for the larger models, while accuracy (AP@0.5) consistently stayed above 0.85. These results confirm that the Jetson can support both lightweight and medium-sized YOLO models with real-time throughput, a stark contrast to the Raspberry Pi.  

This level of throughput enables not only accurate detection but also continuous tracking and the ability to trigger dynamic, responsive deterrence. These results indicate that the full set of evaluated models can be effectively deployed on the NVIDIA Jetson AGX Xavier when combined with efficient preprocessing and post-processing pipelines.

% It is clear that the benefits of VOLTA GPU architecture is significant in reducing latency in NVIDIA Jetson. Using the same ONNX models but with the CUDA execution provider, the Jetson delivered significantly lower inference times across the models. The smallest models, such as YOLOv11n and YOLOv10n, achieved total processing speeds exceeding 40 FPS, well within the requirements for real-time video analysis. Even the medium-sized models, like YOLOv11s and YOLOv8s, maintained performance above 20 FPS. This level of performance enables not just detection but also continuous tracking and the potential for triggering dynamic, responsive deterrence. It means that the entire set of models is deployable on NVIDIA Jetson AGX Xavier with efficient preprocessing and post-processing pipelines.

\subsection{Comparative analysis of model performance}
%Figure \ref{fig:reliability_jetson} and Figure \ref{fig:reliability_pi} demonstrate the trade-off between model reliability (measured by AP@0.5) and latency (measured by inference time) for both the devices in the study. The result makes it easier to identify some models that are more efficient than others. Clearly, n-series models are the fastest across both devices. However, we can observe a subtle, yet contrasting accuracy trend across the devices. 
% Furthermore, results reveal the superior computational efficiency of recent models in comparison to older models. For instance, YOLOv11n (2.58M parameters) achieves a higher accuracy (0.8677 vs. 0.8397 mAP@0.5) without compromising inference speed on the Jetson (68 vs. 76 FPS) compared to the older YOLOv8n (3.01M parameters). This demonstrates that architectural innovations, such as the C2PSA attention mechanism in YOLOv11, are not only theoretical but translate into tangible performance gains on resource-constrained hardware. 

% Similarly, the efficiency of YOLOv10's NMS-free design contributes to its strong performance relative to its size. These findings suggest that for edge deployment, selecting the latest-generation small or medium model (e.g., YOLOv11n/s) is a more effective strategy than using an older, larger model to achieve a similar level of accuracy.

\begin{figure}
    \centering
    \includegraphics[width=0.98\columnwidth]{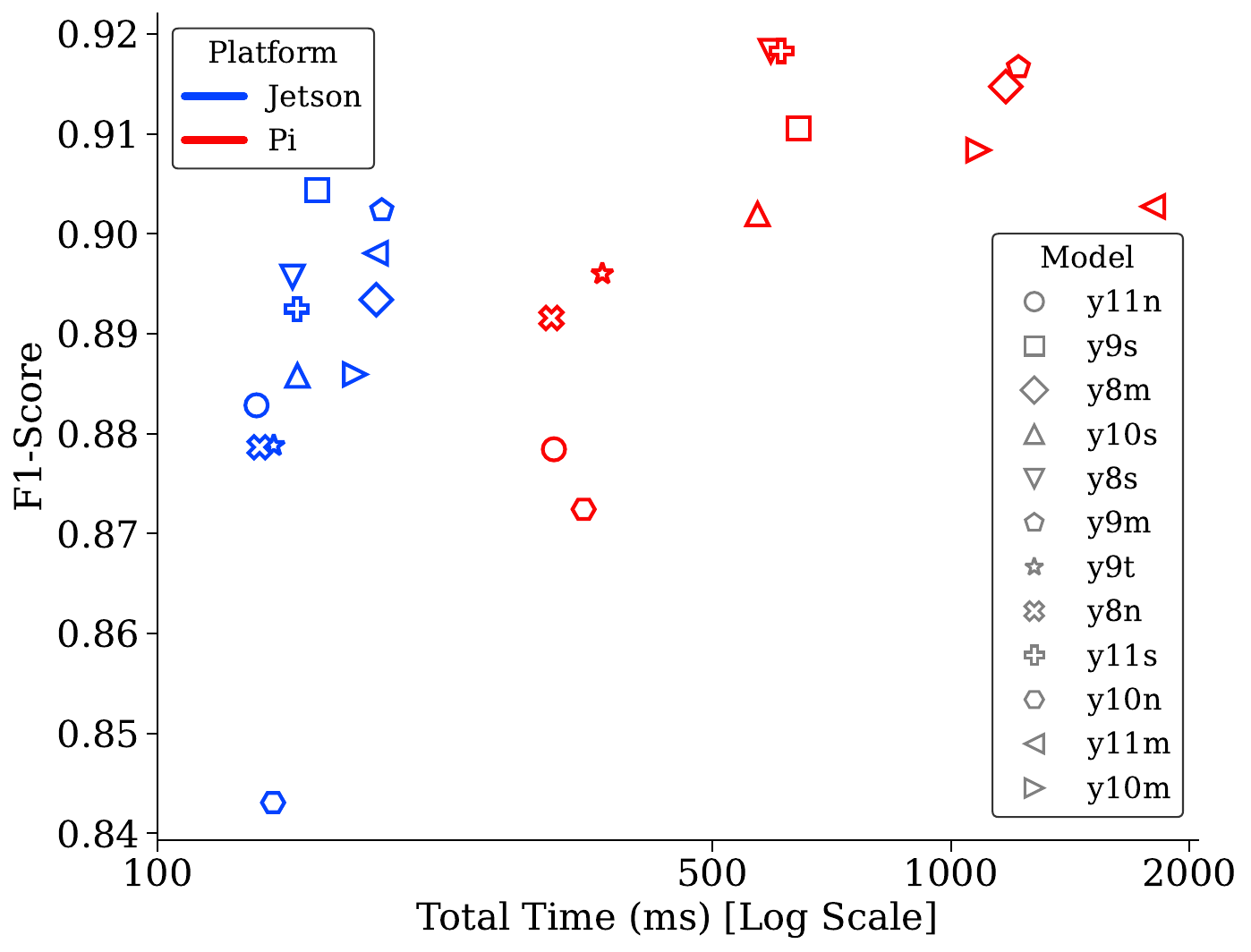}
    \caption{Efficiency frontier in NVIDIA Jetson AGX Xavier (blue) and Raspberry Pi 5 (red).}
    \label{fig:edge_efficiency}
\end{figure}

Figure~\ref{fig:edge_efficiency} provides a comparative evaluation of deer detection models across two edge devices. The time axis is shown on a logarithmic scale to accommodate the orders-of-magnitude difference in processing speed. Two distinct clusters of results are apparent: the NVIDIA Jetson models are concentrated in the low-latency region (below 200 ms), while the Raspberry Pi models are spread across a much wider range, extending from 200 ms to nearly 2000 ms. Here, total time represents the sum of inference, preprocessing, and postprocessing times (see Section~\ref{sec:metrics}). As noted earlier, differences in total time relative to inference time are primarily attributable to software inefficiencies. The plot illustrates the ``efficiency frontier'' for each device, defined as the set of models offering the best accuracy for a given level of performance. On the Jetson, this frontier is nearly vertical, indicating that models can achieve high F1-scores without sacrificing latency. By contrast, the Raspberry Pi’s efficiency frontier is much more extended. Models from the \textit{m} series in all YOLO versions fail to deliver adequate throughput, and even smaller models face latency challenges. Nevertheless, lightweight variants such as YOLOv8n and YOLOv11n achieve the best balance, sustaining competitive accuracy while remaining relatively efficient.  
This analysis highlights that for CPU-bound platforms like the Raspberry Pi, model selection must carefully balance accuracy against the practical frame-rate requirements of real-time applications. In contrast, GPU-accelerated platforms such as the Jetson provide greater flexibility, allowing higher accuracy models to be deployed without significant performance trade-offs.

\subsection{Discussion}\label{sec:discussions}

\subsubsection{Choices of CPU or GPU edge devices}
The choice between CPU- and GPU-based edge devices depends largely on the requirements of the target application. CPU platforms, such as the Raspberry Pi, are low-cost, energy-efficient, and suitable for lightweight tasks where occasional or event-triggered detections are sufficient, for example, confirming the presence of a stationary animal or supporting long-term monitoring in power-constrained environments.
By contrast, GPU-accelerated platforms, such as the NVIDIA Jetson AGX Xavier, are better suited for real-time applications that demand high-frame-rate video analysis, continuous tracking, and rapid decision-making. The Jetson provides the computational capacity to run larger models at interactive speeds, making it ideal for dynamic scenarios where animals are moving, and system responsiveness is critical.
In practice, the Raspberry Pi offers accessibility and low-power operation, while the Jetson delivers the performance necessary for advanced, real-time wildlife detection and deterrence. The decision between the two thus reflects a trade-off between efficiency, cost, and the level of intelligence required at the edge.

% Our extensive model training, evaluation, and deployment compatibility study demonstrated promising results in automatic deer detection using camera images. With the basic YOLO models, without any architectural modification, it is practical to detect deer in wildlife situations with satisfactory accuracy. Even with a small dataset of about 3000 images, YOLO models can perform very well on deer identification, even in the most difficult situations like in low light, distant presence, snowy or rainy environments, and varying lighting and weather conditions. However, there are some limitations to this study as discussed in subsequent sub-sections.

% \subsection{Dependence in GPU}
% The Raspberry Pi is limited to low-frequency, event-triggered snapshots, suitable for confirming the presence of a stationary animal. The Jetson, however, enables true, high-frame-rate video stream analysis, which is a prerequisite for any system intended to interact with or track moving wildlife in real time. This finding suggests that for applications requiring real-time visual intelligence at the edge, investing in a GPU-accelerated platform is not an optimization but a foundational requirement.

\subsubsection{Robustness to other species}
In this work, we curated the Cameratraps dataset, which covers a diverse and challenging range of environmental conditions (see Figure~\ref{fig:dataset}). However, %the dataset is dominated by white-tailed deer, and thus
the ability of the models to generalize to other deer species, or to avoid misclassifying non-deer animals as deer, has not yet been evaluated. For deployment in a real-world automatic deer deterrence system, robustness against such incorrect classifications is essential. Out-of-distribution (OOD) detection methods could be applied to mitigate these risks \citep{liu2021towards}. Achieving this will require additional studies, including the development of a multi-species dataset to rigorously test cross-species generalization and misclassification robustness.  

\subsubsection{Real-world deployment}
Our evaluation on edge devices showed strong detection accuracy across platforms, but real-world deployment also requires speed, efficiency, and reliability. The NVIDIA Jetson AGX Xavier offered a favorable balance of performance and accuracy, supporting real-time operation. In contrast, smaller low-power devices such as the Raspberry Pi 5 performed poorly with larger models, where high GFLOPs demands quickly overwhelmed the CPU. The study further indicates that simply converting PyTorch models to ONNX and running them with ONNX Runtime is insufficient for achieving real-time performance on the Raspberry Pi. Additional optimizations are necessary, including lightweight frameworks such as TensorFlow Lite and improvements in preprocessing and postprocessing pipelines \citep{iqbal2024review}. To enable real-time deployment on constrained platforms like the Raspberry Pi 5, future efforts should focus on advanced model optimization techniques, more efficient inference engines, and lighter architectures. These measures are essential to minimize resource usage while maintaining acceptable detection performance in practical field settings.

\section{Conclusion}\label{sec:conclusion}
In this paper, we introduced an open-source dataset of 3,095 annotated deer images with bounding-box labels, capturing diverse environmental conditions and lighting scenarios. We benchmarked 12 model variants from four recent YOLO architectures (v8, v9, v10, and v11) and evaluated their performance on two representative edge devices: the CPU-based Raspberry Pi 5 and the GPU-powered NVIDIA Jetson AGX Xavier. Results showed that larger m-series models achieved the highest accuracy on high-end hardware, with AP@0.5 scores exceeding 0.94. However, their computational demands made them unsuitable for real-time deployment on CPU-only devices such as the Raspberry Pi 5, where performance dropped below 1 FPS. In contrast, the Jetson AGX Xavier provided an optimal balance, sustaining real-time processing speeds above 25 FPS while maintaining high detection accuracy (AP@0.5 > 0.85). These findings demonstrate that GPU-accelerated hardware is a prerequisite for real-time wildlife tracking at the edge. Overall, this study provides clear, actionable guidance for the design of effective autonomous deer detection systems that can be deployed on edge devices. Since fast and accurate deer detection is fundamental to advanced monitoring, tracking, and deterrence applications, this study plays a foundational role in the development of such systems.

Future work will focus on hardware-specific optimization techniques and lightweight frameworks, such as TensorFlow Lite, to further improve performance on constrained devices. Additionally, we plan to expand the dataset by collecting and annotating images that capture a wider range of challenging conditions, including adverse weather (e.g., snow, heavy rain, fog), diverse agricultural landscapes (e.g., cornfields, soybean fields), and a greater variety of deer species and behaviors.

\section*{Authorship Contribution}
\textbf{Bishal Adhikari}: Conceptualization, Investigation, Software, Writing – Original Draft;
\textbf{Jiajia Li}: Conceptualization, Writing – Original Draft; 
\textbf{Eric S. Michel}: Conceptualization, Writing - Review \& Editing; 
\textbf{Jacob Dykes}: Conceptualization, Writing - Review \& Editing;  \textbf{Te-Ming Paul Tseng}: Conceptualization, Writing - Review \& Editing; 
\textbf{Mary Love Tagert}: Conceptualization, Writing - Review \& Editing; 
\textbf{Dong Chen}: Conceptualization, Investigation, Supervision, Writing - Review \& Editing.

\section*{Acknowledgement}
This publication is a contribution of the Forest and Wildlife Research Center, Mississippi State University, and this material is based upon work that is supported by McIntire-Stennis funds under accession number (7010294). We also thank the Idaho Department of Fish and Game for making the Idaho Camera Traps dataset \citep{idaho_cameratraps} publicly available for research. 

\typeout{}
\bibliography{ref}
\end{document}